\documentclass{article}
\usepackage{ijcai19}

\usepackage{times}
\usepackage{soul}
\usepackage{url}
\usepackage[hidelinks]{hyperref}
\usepackage[utf8]{inputenc}
\usepackage[small]{caption}
\usepackage{graphicx}
\usepackage{amsmath}
\usepackage{booktabs}
\usepackage{algorithm}
\usepackage{algorithmic}

\usepackage{amsfonts}
\usepackage{subfigure}

\title{MiniMax Entropy Network: Learning Category-Invariant Features\\ for Domain Adaptation}

\author{
	Chaofan Tao$^1$\and
	Fengmao Lv$^1$\and
	Lixin Duan$^1$\footnote{Contact Author}\And
	Min Wu$^2$\\
	\affiliations
	$^1$Big Data Research Center, University of Electronic Science and Technology of
	China\\
	$^2$Institute for Infocomm Research, A*STAR\\
	\emails
	\{tcftrees, lxduan\}@gmail.com,
	fengmaolv@126.com,
	wumin@i2r.a-star.edu.cn
}

\begin{document}
	
	\maketitle
	
	\begin{abstract}
		How to effectively learn from unlabeled data from the target domain is crucial for domain adaptation, as it helps reduce the large performance gap due to domain shift or distribution change. In this paper, we propose an easy-to-implement method dubbed MiniMax Entropy Networks (\textit{MMEN}) based on adversarial learning. Unlike most existing approaches which employ a generator to deal with domain difference, MMEN focuses on learning the categorical information from unlabeled target samples with the help of labeled source samples. Specifically, we sets  an \textit{unfair} multi-class classifier named category discriminator, which classifies source samples accurately but be confused about the categories of target samples. The generator learns a common subspace that aligns the unlabeled samples based on the target pseudo-labels. For MMEN, we also provide theoretical explanations to show that the learning of feature alignment reduces domain mismatch at the category level. Experimental results on various benchmark datasets demonstrate the effectiveness of our  method over existing state-of-the-art baselines.
		%from the source labels as well as pseudo labels inferred for unlabeled target samples at each training iteration.
	\end{abstract}
	
	\section{Introduction}
	Though deep convolutional networks has gained great advancement in visual understanding over the past years \cite{bin2019mr,tao2020dynamic,chen2021litegt,tao2021fat}, its training process heavily relies on numerous labeled samples. Since it is often prohibitively expensive to manually label a large-scale  dataset for a certain learning task at hand, how to effectively relieve the annotation burden in deep learning remains an open issue.
	
	In recent years, synthetic images, whose class labels can be cheaply generated with the recent advances of computer graphics techniques, are tentatively used to train  models that can work in real-world scenarios, aiming to reduce the corresponding labelling consumption \cite{peng2017visda}.  However, the domain discrepancy between the synthetic images (i.e. source domain) and the real-world photos (i.e.  target domain) still severely degrades the performance of model. As theoretically discussed in \cite{ben2010theory}, this discrepancy can lead to statistically unbounded risk for target tasks. To improve the model's generalization ability across domains, domain adaptation has been widely studied especially for unsupervised domain adaptation.
	
\begin{figure}[t]
		\centering
		\includegraphics[width=0.90\linewidth]{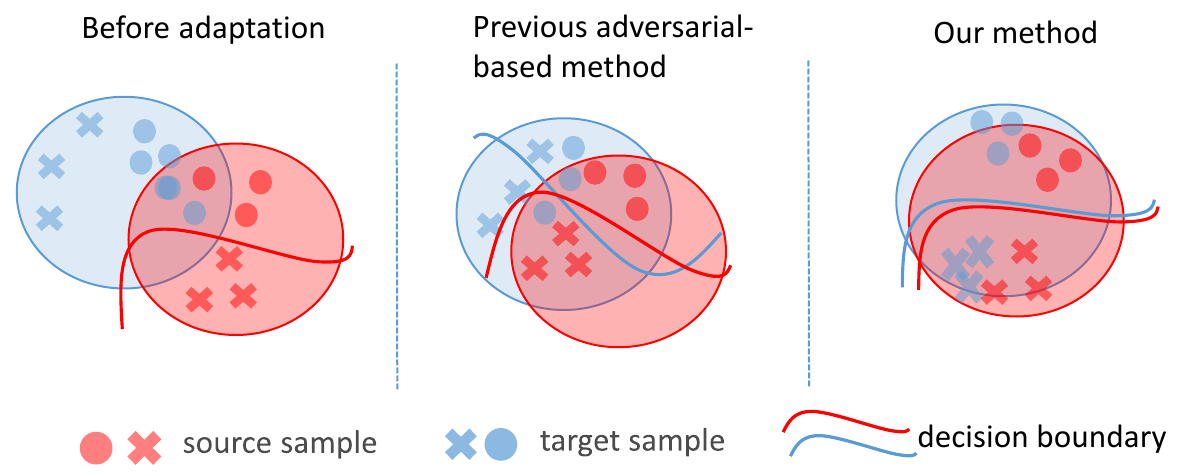}
		\caption{A comparison of feature alignment between previous adversarial-based method and our method. Red line and blue line denote the decision boundary for source domain and target domain respectively. Different shapes indicates different categories. \textbf{Left:} the features are scattered irreuglarly before adaptation, and the source samples can be classified accurately with available source labels. \textbf{Middle:} Previous adversarial-based methods employ domain classifiers to achieve domain confusion, thereby aligning the feature distribution globally in the level of domain. However, domain classifier contains no categorical information, therefore category shift between domains is ignored. \textbf{Right:} Our model utilizes an \textit{unfair} category discriminator to align the target feature according to the decision boundary for the source, Threfore our model enables category-level feature alignment.}
		\label{figure0}
	\end{figure}

	Formally, unsupervised domain adaptation aims to achieve desirable results in the target domain through adaptation from  labeled source dataset and unlabeled target dataset \cite{pan2010survey}. As indicated above, the distribution discrepancy forms the main bottleneck in domain adaptation.  In order to learn a transfer model across domains, various approaches has been proposed including discrepancy-based methods \cite{long2015learning,long2016unsupervised,bousmalis2016domain}, reconstruction-based methods \cite{bousmalis2016domain,Gen2Adapt} and adversarial-based methods \cite{ganin2014unsupervised,tzeng2017adversarial}. Specifically, adversarial-based methods  are widely implemented in recent years. As shown in Figure \ref{figure0}, Most of existing adversarial-based methods set a domain classifier as discriminator to judge the origin of input samples, thereby pushing the model to generate domain-invariant feature representations. However, we argue that two issues exist in the existing works: (1) In classification tasks, the data distribution over intermediate representations are mixture of different clusters, but  classic adversarial-based methods cannot precisely align the clusters with each specific category since domain labels have no categorical information ; (2) To achieve good generalization performance over the target domain, the target samples need to be kept far way from the decision boundaries, but domain invariance can only account for global alignment. Therefore, it is essential to consider categorical information of target images during the learning process.

	In this paper, we propose a novel adversarial training model dubbed MiniMax Entropy network (MMEN), which is implemented in an end-to-end fashion. The architecture of our model can be viewed as a variant of Domain-Adversarial Neural Networks (DANN) \cite{ganin2014unsupervised}. Instead of using a domain classifier as discriminator, we set an \textit{unfair} multi-class classifier named category discriminator to incorporate category information into the adversarial procedure. As displayed in Figure \ref{figure2}, our model contains a shared feature generator $G$, a category discriminator $D$ and an auxiliary classifier $C$. The adversarial procedure is implemented through controlling the information entropy of the discriminator's softmax predictions over target samples. It is noteworthy that this entropy quantity can reflect the discriminator's confidence in the target samples' label assignment \cite{krause2010discriminative}. During the training process, $D$ is trained to classify the source images confidently but be extremely confused about the category of the target images. In contrast, $G$ aims to generate features that assist the target samples to be classified by $D$ with high confidence. With the support of the source domain, $D$ is able to detect target samples that are not aligned with the source features of any categories, while $G$ will be guided to generate target features that mimic the source feature of each specific category. Through the minimax game between $G$ and $D$, our model can achieve domain invariance within each category, which can lead to discriminative features for the target domain.
	
	Overall, the contributions of this work are listed as follows:
	\begin{itemize}
		\item  We propose MMEN for unsupervised domain adaptation.  MMEN is trained through a minimax game over the entropy of the category discriminator's prediction, which encourages to achieve domain invariance within each category.  
		%Our model is capable of generating feature representations that enjoy invariance  with the shift among categories between domains.
		\item Our model  enjoys a concise framework and a clear training procedure. Therefore, our model is easy-to-implement and efficient compared with other methods.
		\item The experimental results outperform the existing state-of-the-art methods on various  benchmark datasets. Then we make a completed evaluation.
	\end{itemize}
	
	\section{Related Work}
	
	\subsection{Unsupervised Domain Adaptation}
	How to perform unsupervised domain adaptation remains a open issue theoretically and practically. Since deep models tend to generate more transferable and informative features than the shallow models \cite{donahue2014decaf}, the recent domain adaptation methods primarily focus on achieving domain invariance in intermediate layers of convolutional neural networks. Specifically, methods that utilized maximum mean discrepancy (MMD) are representative, including \cite{long2015learning,long2016unsupervised,bousmalis2016domain,long2017deep} that align the distributions between different domains by subspace learning.  \cite{long2015learning} proposed to reduce the domain  discrepancy through embedding the intermediate features into reproducing kernel Hilbert space and then minimizing the means of data distribution  between domains. In \cite{long2017deep}, the MMD measurement was further implemented over the joint distributions of features and labels, aiming to correct both domain shift \cite{pan2010survey} and conditional shift \cite{zhang2013domain}.
	
	Besides MMD, adversarial-based methods for domain adaptation have sprung up recently with appearance of the generative adversarial networks (GAN) \cite{goodfellow2014generative}. In order to effectively bridge source domain and target domain, the previous methods \cite{ganin2014unsupervised,tzeng2017adversarial,chadha2018improving} primarily focused on learning domain-invariant representations, by which the distribution discrepancy between domains can be reduced. These methods utilized a domain classifier to achieve domain alignment in DNNs \cite{ganin2014unsupervised}. The domain classifier was trained to distinguish the origin of input based on feature representations, while the generator aimed to cheat the discriminator by generating features with small domain discrepancy. However, domain invariance does not necessarily imply discriminative features for target data.
	
	Among the most recent approaches, diverse adversarial-based methods attempted to inject task knowledge to their models indirectly. In \cite{chen2018re}, the author used target samples to alleviate the domain shift by re-weighting the distribution of source labels.  In \cite{pei2018multi}, the target pseudo-labels were employed to weight the loss on multiple subordinate domain classifiers. In \cite{zhang2018collaborative}, selected target pseudo-labels were combined with learnable weight functions in the classification loss.  Aforementioned methods ignore the feature distribution alignment for each specific category, which  more or less degrade the model's generalization ability over the target domain.
	
	\subsection{Loss Regularization}
	Loss regularization has been widely applied to benefit parameterized models of posterior probabilities from unlabeled data or partially labeled data\cite{grandvalet2006entropy}.  \cite{krause2010discriminative} proposed to estimate the data distribution by entropy for clustering data and training a classifier simultaneously.  \cite{huang2022frequency} proposes a loss function for improving adversarial robustness. \cite{springenberg2015unsupervised} modeled a discriminative classifier from unlabeled data by maximizing the mutual information between inputs and predicted categories. In the field of domain adaptation,  \cite{tzeng2015simultaneous} once employed cross-entropy between target activation and soft labels to exploit semantic relationships in label space.
	
		\begin{figure*}[t]
		\centering
		\includegraphics[width=0.95\linewidth]{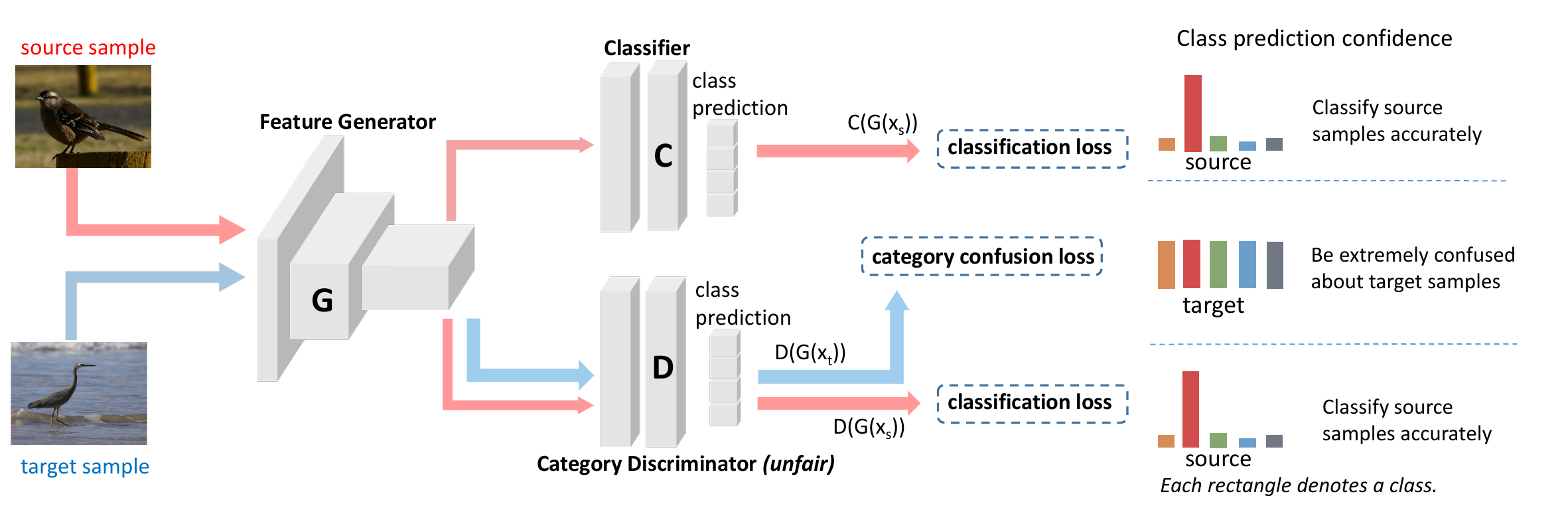}
		\caption{An overview of our model during the learning process. Red line and blue line denote source flow and target flow respectively. The category discriminator D is an \textit{unfair} multi-class classifier that classify source samles accurately while being confused about target samples.The generated target pseudo-labels $D(G(x_t))$ are directly utilized in adversarial training for feature alignment. By maximizing the category confusion on the target samples through D, the generators (G) are guided to align target features alike the source ones for high confidence. The unfairness design effectively transfers the source discriminability to the target domain. Besides, we apply classification loss to an auxiliary classifier C to enhance the  categorical discriminability towards source samples}
		\label{figure2}
	\end{figure*}
	\section{Method}
	Consider $\mathcal{X}$ as the input space and $\mathcal{Y}$ as the label space, source data $x_s$ and target data $x_t$ are drawn from marginal distribution $P(X_s)$ and $ P(X_t)$ respectively. We use $\left \{ (x_s,y_s) \right \} $ as  source dataset, which belongs to source domain $D_s$. Then we use $\left \{ x_t \right \} $ as unlabeled target dataset, which belongs to target domain $D_t$.  It is assumed that the data in different domains share the feature space while following different marginal distributions $P(X_s) \neq P(X_t)$, named as domain shift. The aim of unsupervised domain adaptation is to train a flexible model $\eta :\mathcal{X}\rightarrow \mathcal{Y}$ that has low target risk $\Pr_{(x_t,y_t)\sim D_t}[\eta(x_t)\neq y_t]$ using all given data.
	\subsection{Motivation}
	% 1. problem description and improve the model from what aspects
	% 2. the function of categorical D and entropy utilization
	
	%From the level of domain, the min-max game between generator and discriminator can help generate domain-invariant representation assuming these models reach a equilibrium state.
	
	Most recent adversarial-based method treat the discriminator as a domain classifier that discriminates the domain of input. 
	If we take a close look, the generator could not receive any target categorical information from the domain classifier during the learning process, thus the generator has problem aligning feature distribution in the level of category.  Accordingly, these models could not guarantee that the generated features were categorically discriminative.  For any specific category, the target features that far away from the source ones are likely to be classified into wrong class even with high confidence. To tackle this problem, we propose a method that utilize categorical information to help the generator to generate category-invariant feature.
	
	Since the source data is labeled, a well-trained classifier $C$ for source data can be learned easily with convolution neural network. Target risk is potential to be reduced significantly with effective alignment.  However, it is challenging to find a reasonable way for the classifier to score the prediction of target samples, since we have no access to the target labels. ``Scoring'' can be viewed as a soft label assignment. We find that soft label assignment can be achieved simply and efficiently by utilizing iteratively-updating pseudo-labels in an adversarial manner.
	% We explain the role of pseudo-labels in the adversarial learning in terms of information entropy, and provide theoretical analysis in terms of GAN-based theory.
	% Previous methods solved it by either using the average activation values on the categories in the source only model instead of target one-hot labels~\cite{tzeng2015simultaneous}, or data augmentation~\cite{motiian2017few}. However, these methods suffer from two issues: (1) Fixed source only model cannot provides accurate predictions on target samples, thus injecting unneglectable noise and making the loss function that related to target samples fail to convergence.  (2) Aforementioned methods are not workable in an unsupervised manner. 
	
	In our model, we set a category discriminator $D$ after the generator.  The discriminator assigns the feature representation of each target sample to a probabilistic vector. $D$ is a muli-class predictor in this paper, but undertakes different responsibilities in different steps.  $D$ is first pre-trained on the source samples to ensure the model obtain categorically discriminative representations of source and then make accurate predictions. During the step of category confusion, $D$ is streamed with both source samples and target samples, aiming to push the generator to align feature distribution in the level of category.  Our method does not require to set domain labels for adversarial training.

	\subsection{Model Description}
%		\subsubsection{Domain Confusion}	
%		In order to accomplish the aforementioned goals for classifier and discriminator, we first pre-train the model by classifying the source samples with available labels. We set the cross-entropy between true labels and probabilistic outputs as the classification loss, which is defined as follow:
%		\begin{equation}
%		\ L_c=   -\frac{1}{n_s} \sum_{x_s} [  y_s \cdot \log(C(G(x_s)))],
%		\label{e00}
%		\end{equation}
%		where $n_s$ denotes the number of samples in source domain in a batch. Dot product is used as ``$\cdot$''. It is an essential step for the model to take measure on target samples because the source only model can give plausible support on target samples at the start of adaptation. 
%		
%		The concept of domain confusion is first proposed in \cite{ganin2014unsupervised}. In this work, a domain discriminator is employed to play minimax game with a feature extractor. To achieve domain confusion, the domain confusion loss $L_{dc}$ is defined as follow:
%	\begin{equation}
%L_{dc} =   - \frac{\lambda_0}{n_s+n_t}\!\!\!\!\! \sum_{x \in D_s \cup D_t} [d \cdot \log (D(G(x)))],
%	\label{e01}
%	\end{equation}
%	where  $d \in \{0,1\}$ denote the domain labels. $\lambda_0$ is a trade-off parameter. By training the network as follow:
%	\begin{equation}
%		\mathop {\min } \limits_{G,C} \mathop {\max } \limits_{D}  \ \ L_c -L_{dc},
%			\label{e02}
%	\end{equation}
%	 the  domain shift can be reduced  globally with learning domain-invariant representations, while category difference between domains is igonred.

	 \subsubsection{Category Confusion}
	%%%%%%%%% figure %%%%%%%%%%

		In order to accomplish the aforementioned goals for classifier and discriminator, we first pre-train the model by classifying the source samples with available labels. We set the cross-entropy between true labels and probabilistic outputs as the classification loss, which is defined as follow:
		\begin{equation}
	\begin{aligned}
\mathop {\min } \limits_{G,C,D} \ L_c=   -\frac{1}{2n_s} \sum_{x_s} [ & y_s \cdot \log(C(G(x_s))) \\
+ &  y_s \cdot \log(D(G(x_s)))  ],
\end{aligned}
		\label{e00}
		\end{equation}
		where $n_s$ denotes the number of samples in source domain in a batch. Dot product is used as ``$\cdot$''. It is an essential step for the model to take measure on target samples because the source only model can give plausible support on target samples at the start of adaptation. 

 The goal of our model is to jointly train the categorical discriminator $C$ and feature extractor $G$ and  to generate category-invariant feature. As shown in the Figure \ref{figure2}, $C$ and $G$ hold different stands:
	\begin{itemize}
		\item \textbf{(D) Category Discriminator’s stand} Classify source samples accurately; Be uncertain of the predicted categories on target samples.
		\item \textbf{(G) Feature Generator’s stand} Generate category-invariant feature representations.
	\end{itemize}
	
	\noindent The discriminator need to judge the domain of input based on feature distribution,  then classify source samples accurately and make \textit{unfair} predictions on target samples. \textit{Unfair} means the uncertainty about predicted category. The generator strives to align target representations that mistake the discriminator for the source to enjoy accurate predictions. If the distribution of generative target feature $P_g$ is consistent with the distribution of source feature $P_f$ in the level of both domain and category, the classifier can predict target samples correctly with high confidence.
	
	In order to achieve uncertainty of the predicted categories, we need to estimate the degree of error predictions on target samples by discriminator, and feedback this information to the generator. Although the labels of target samples are inaccessible, we can utilize pseudo-labels $\hat{y_t} = D(G(x_t))$. A nature way to estimate the error predictions is to judge whether the cross-entropy of pseudo-labels $H(p(y|x_t))$ is sufficiently small, since it is minimized when the discriminator make certain predictions. Alternatively speaking, $H(p(y|x_t))$ is maximized when the distribution of class prediction is even (extremely confused). The cross-entropy of pseudo-labels $H(p(y|x_t))$ can be formalized as follow:
	\begin{equation}
	\begin{aligned}
	H(p(y|x_t)) =  -\frac{1}{n_t} \sum_{x_t} \hat{y_{t}}\cdot \log (D(G(x_t))),
	\end{aligned}
	\label{e2}
	\end{equation}
	% H(p(y|x_t)) = -\frac{1}{n_t} \sum_{x_t} \hat{y_{t}}\cdot \log (D(G(x_t))),
	\noindent where $n_t$ denotes the number of samples in target domain in a batch. Assuming that there are $K$ categories in the dataset, then $\hat{y_{t}}$ is a $K$-dimensional vector denoting class-wise probability of a target sample,  $\hat{y_{t}}^{(i)}\!= p(y_i\!\!=\!\!1|x_t)$ for $i=1,\cdots,K$. 
 \\
	\indent	Remember that we want the target feature to mimic the source distribution, thereby utilizing the capacity of source model to infer the target labels. As we know,	cross-entropy reflects the relationship between feature distribution and decision boundary. High cross-entropy of pseudo-labels $	H(p(y|x_t)) $ means that the generated target features are near the decision boundaries of category discriminator, thereby making the discriminator confused about the predicted categories. During the process of  training, the category discriminator is expected to keep unfair. This setting guides the generator to generate  category-invariant feature for high prediction certainty until the generated target features are fully aligned. Therefore, we  maximize $	H(p(y|x_t)) $ for the D, which tends to assign target pseudo-labels that have equal probability to each category.  By minimizing $	H(p(y|x_t)) $ for the G, it tends to generate source-alike feature for the target sample.  Our final objective is:
%	\indent To further exploit the category information during training, we want our model to pay different attentions aligning different target samples according to their \textit{difficulty} .   In this paper, the \textit{difficulty} denotes the degree of alignment. For the hard samples that are weakly aligned, the weight of alignment loss is supposed to be large. By contrast, the easy samples just need slight alignment towards source feature. Inspired by the domain discriminator, we employ the domain prediction to adaptively weigh the difficulties of target samples. The domain labels are set as 0 and 1 in source domain and target domain, respectively. The target samples are considered as hard if their domain predictions $D(G(x_t))$ are large, correspondingly we increase the weight of alignment-related loss function. In our model, we introduce the cross entropy of pseudo-labels for feature alignment. Hence, the final category confusion loss $L_{cc}$ is defined as follow:
%	\begin{equation}
%	L_{cc} =  -\frac{\lambda}{n_t} \sum_{x_t}   \hat{y_{t}}\cdot (D(G(x_t))\log (C(G(x_t)))),
%	\label{e3}
%	\end{equation}
			%%%%%%%%% figure %%%%%%%%%%
%	\begin{figure*}[t]
%		\centering
%		\includegraphics[width=1.0\linewidth]{align.pdf}
%		\caption{Illustration of feature alignment in the level of both domain and category. Different colors denote different categories.}
%		\label{figure2}
%	\end{figure*}
	%%%%%%%%% figure %%%%%%%%%%
	 
	\begin{equation}
		\mathop {\min } \limits_{G} \mathop {\max } \limits_{D}  \ \  \lambda	H(p(y|x_t)),
\label{e4-1}
	\end{equation}	
		\begin{equation}
	\mathop {\min } \limits_{G,D,C}   \ \ L_c,
\label{e4-2}	
	\end{equation}
where $\lambda $ controls the trade-off between classification and category confusion. 
	\subsubsection{Training Procedure}

    First, in order to make the classifier keep discriminative on source, we apply apply Eq.\ref{e00} to  train the source only network. Second, we use Eq.\ref{e4-1} and Eq.\ref{e4-2} to learn the generator G ,  the unfair category discriminator D and classifier C. Notice that  classifier C is auxiliary used to  enhance the  categorical discriminability towards source samples. Our model  works fine without C, we will discuss this phenomenon later.
    
    It is worthnotey that misuse of the noisy pseudo labels could have negative effects during the training procedure empirically. That is just the one that motivates us to leverage the pseudo labels through the mini-max game over entropies, but not directly assigning pseudo labels to target data. As the learning process develops, the aligned target feature will cause the cross entropy of pseudo labels decreases gradually, thereby generating relatively clean pseudo labels. The performace will be improved in cycle. We discuss that our model is capable of inferring target labels successfully step by step. Notice that our model does not need to filter bad target samples  manually like setting an accuracy threshold. All target samples can be used during the training process.
    
    Since C and G hold opposite stands on target samples, we optimize the model by iterative training. To balance the power of both sides, we set a hyper-parameter $k$, which signifies the times of updating G before updating C in a batch. In the test phase, target samples are fed forward through G and C for final prediction.

	\section{Experiments}

% 	\begin{table*}[t]
% 		\centering
% 		\renewcommand\tabcolsep{6pt}
% 		\begin{tabular}{lrrrrrrrrrrrrr}
% 			\toprule
% 			\textbf{Model}  & plane & bcycl &  bus  &  car & horse & knife & mcycl & person & plant & sktbrd & train & truck & Avg.  \\
% 			\midrule
% 			Source Only         &55.1 &53.3 &61.9 &59.1 &80.6 &17.9 &79.7 &31.2 &81.0 &26.5 &73.5 &8.5 &52.4 \\
% 			DANN                &81.9 &\textbf{77.7} &\textbf{82.8} &44.3 &81.2 &29.5 &65.1 &28.6 &51.9 &54.6 &82.8 &7.8 &57.4 \\
% 			\textbf{MMEN}      &\textbf{83.1} &27.3 &60.6 &\textbf{74.7} &\textbf{83.6} &\textbf{34.1} &\textbf{94.8} &\textbf{50.3} &\textbf{83.9} &\textbf{60.6} &\textbf{83.8} &\textbf{16.4} &\textbf{62.8} \\
			
% 			\bottomrule
% 		\end{tabular}
% 		\caption{Comparison of the performance (\%) for unsupervised domain adaptation on the VisDA dataset.}
% 		\label{table1}
% 	\end{table*}

	 To evaluate the proposed method on several benchmark datasets, we employ classification accuracy as metric. The metric is define as follow:
	 \begin{equation}
	     Accuracy = \frac{|x_t:x_t \in X_t \ \wedge \ \hat{y_t}=y_t|}{|x_t:x_t \in X_t|}.
	     \label{e8}
	 \end{equation}
    All our experiments are implemented by PyTorch. We set the hyper-parameter $k=4$ and coefficient $\lambda=0.1$ in all the experiments. Results show the superiority of our method against state-of-the-art transfer learning methods.

	\subsection{ImageCLEF-DA Dataset}
	ImageCLEF-DA\footnote{\url{http://imageclef.org/2014/adaptation}} is a benchmark dataset for domain adaptation. We use 3 domains of data in the dataset, including ImageNet ILSVRC 2012 (I), Pascal VOC 2012 (P) and Caltech-256 (C).  Each domain has 12 shared categories and each category has 50 images. We perform all transfer tasks  across domains, namely  I $\rightarrow$ P, P $\rightarrow$ I, I $\rightarrow$ C, C $\rightarrow$ I, C $\rightarrow$ P and P $\rightarrow$ C.
	
	We utilize ResNet50~\cite{he2016deep} as our CNN architecture.  For fair comparison, we compare the performance of MMEN with methods that based on ResNet50.  Both classifier and category discriminator are three fully-connected layers (1000$\rightarrow$1000$\rightarrow$12) with batch normalization layers. The model is pre-trained based on ImageNet and then fine-tuned by source samples initially. SGD is used as the optimizer with learning rate $1\times 10^{-3}$. The batch size of both source samples and target samples are 24 equally. We train our model 150 epoches in total.
	
% 	\subsection{VisDA Dataset}
% 	VisDA ~\cite{peng2017visda} is a new large-scaled dataset for \textit{the 2017 Visual Domain Adaptation challenge} aiming to solve the real-world problem of domain shift. The dataset contains 152,397 synthetic images for training  and 55,388 real-world  images  for validation. Training dataset is used as source domain and validation dataset is used as target domain in our experiments.  There are 12 common categories of data in each domain. We conduct experiments based on pre-trained ResNet101 and compare the results with other methods that used the same architecture. We employ the same hyper-parameter implemented on ImageCLEF-DA dataset except that the batch size is 32.

	\subsection{Digits Dataset}
	MNIST~\cite{lecun1998gradient}, SVHN~\cite{netzer2011reading}, and USPS~\cite{hull1994database} , which consist 10 classes of numbers, are three commonly used datasets in digit classification and domain adaptation.  We perform three challenging transfer tasks  SV $\rightarrow$ MN,  MN $\rightarrow$ US and  US  $\rightarrow$ MN to evaluate our method.
	
	We follow the shallow CNN architecture used in~\cite{ganin2014unsupervised}. Batch normalization layers are employed on each layer. We use Adam~\cite{kingma2014adam} as the optimizer and learning rate is $2\times 10^{-4}$. Batch size is set as 128 for both source samples and target samples.  We train our model 150 epoches totally
	
			\begin{table}[t]
		\centering
		\renewcommand\tabcolsep{3.0pt}
		\begin{tabular}{lrrrrrrr}
			\toprule
			\textbf{Model}  & I \!\!\! $\rightarrow$ \!\!P & P \!\!$\rightarrow$ \!\!I & I\!\! $\rightarrow$ \!\!C & C\!\! $\rightarrow$ \!\!I & C\!\! $\rightarrow$ \!\!P & P\!\! $\rightarrow$\!\! C &\!\! Avg. \\
			\midrule
			ResNet50          &74.8	        &83.9	&91.5	              &78.0	&65.5	&91.2	&80.7 \\
			DAN               &74.5         &82.2   &92.8                 &86.3 &69.2  &89.8    &82.5 \\
			RTN               &74.6         &85.8   &94.3                 &85.9 &71.7  &91.2    &83.9 \\
			DANN              &75.0	        &86.0	&\textbf{96.2}	      &87.0	&74.3	&91.5	&85.0 \\
			JAN               &76.8	        &88.0	&94.7	              &89.5	&74.2	&91.7	&85.8 \\
			MADA              &75.0	        &87.9	&96.0	              &88.8	&75.2	&92.2	&85.8 \\
			%CAN	              &78.2	        &87.5	&94.2	              &89.5	&75.8	&89.2	&85.7 \\
			%iCAN              &\textbf{79.5}&89.7	&94.7	              &89.9	&\textbf{78.5}	&92.0	&87.4 \\
\midrule
				\textbf{G+D} 	&76.2	&91.3&	94.8&	89.1&	72.9&	94.0	&86.4 \\
				
			%	\textbf{D+C}  &  77.3 &  90.2 & 	94.2 & 	89.0 & 	73.0 & 	93.8 & 86.3 \\	
			\textbf{MMEN }       & \textbf{77.8} &\textbf{92.2} &95.8        & \textbf{89.8} & \textbf{75.8} & \textbf{94.5} & \textbf{87.7} \\
	\midrule
				Oracle & 95.5	&99.7	&100.0&	99.7&	95.5&	100.0 & 98.4\\
			\bottomrule
		\end{tabular}
		\caption{Comparison of the performance (\%) for unsupervised domain adaptation on the ImageCLEF-DA dataset. Oracle denotes the target samples are trained in a fully supervised manner.}\label{table2}
	\end{table} 
	
	\subsection{Comparative Results}
	To verify the effectiveness of each module in our model, we consider several variants as follow: 
	\begin{itemize}
	\item{\textbf{G+D}} \  We utilizze the target pseudo labels to adversarily train the generator G and category discriminator D. 
	%\item{\textbf{D+C}} We first employ the domain confusion to train the feature extractor G and domain discriminator D, and then use category confusion loss to train G and C.
	\item{\textbf{MMEN}} \ Compared with `G+D', we add an auxiliary classifier C during training. It is the full model of minimax entropy network, namely MMEN.
	\end{itemize}
	We compare our results with the state-of-the-art methods, including
	\textbf{ResNet50}~\cite{he2016deep},
	Deep Adaptation Network (\textbf{DAN}) ~\cite{long2015learning} ,
	Residual Transfer Network (\textbf{RTN}) ~\cite{long2016unsupervised},
	Domain separation networks (\textbf{DSN}) ~\cite{bousmalis2016domain},
	Joint Adaptation Network (\textbf{JAN}) ~\cite{long2017deep},
	Domain-Adversarial Neural Networks \textbf{(DANN)} ~\cite{ganin2014unsupervised} ,
	Multi-Adversarial Domain Adaptation (\textbf{MADA}) ~\cite{pei2018multi},
	Re-weighted Adversarial Adaptation Network (\textbf{RAAN}) ~\cite{chen2018re},
	Adversarial Discriminative Domain Adaptation (\textbf{ADDA}) ~\cite{tzeng2017adversarial} and its improved variant (\textbf{iADDA}) ~\cite{chadha2018improving}.
	%Collaborative and Adversarial Network (\textbf{CAN})  ~\cite{DBLP:conf/cvpr/ZhangO0018} and its variant (\textbf{iCAN}).

	As shown on Table~\ref{table2} and Table~\ref{table3}, we can observe the comparative results. MMD-based methods DAN, RTN , DSN and JAN  exhibit relatively low performance due to coarse-grained feature alignment. Compared with the adversarial-based methods DANN and ADDA that make adaptation on generator without any categorical information, our method improves the performance with a remarkable gap. The most recent methods MADA, RAAN and iADDA utilize task knowledge indirectly by re-weighting the loss function or mimicking fixed source encoder posteriors. By contrast, our model injects target knowledge directly and enables fine-grained feature alignment, therefore it outperforms state-of-the-art methods on various benchmark datasets. The easy-to-implement framework and competitive performance indicates that play minimax game over pseduo labels is an effective and efficient way for unsupervised domain adaptation.	The auxiliary classifier C in MMEN enhance  discriminability towards source samples. The module help stablize the distribution that the target feature aims to align. Hence, MMEN obtains better performace compared with `G+D'.

	\begin{table}[t]
		\centering
		\renewcommand\tabcolsep{12.5pt}
		\begin{tabular}{lrrrr}
			\toprule
			\textbf{Model}  & S\! $\rightarrow$\! M &  M\! $\rightarrow$\! U &  U \!\! $\rightarrow$\! M     \\
			\midrule
			Source Only         &67.1      &76.7    &63.4     \\
			DANN                &73.9   & 77.5   &-    \\
			DSN                 &82.7   & -   & -       \\
			ADDA                &76.0   & 89.4   & 90.1  \\
			RAAN                &89.2  &89.0  &92.1     \\
			iADDA               &92.7   & 91.0  & 94.8 \\
			%MCD                 &96.2   & 94.2  & 94.1  & 94.4 \\
		\midrule
			\textbf{G+D}  & 97.2 & 96.8 &96.6 \\
			%\textbf{D+C} \\
			\textbf{MMEN}      & \textbf{98.8}	& \textbf{97.8}	& \textbf{97.4}  \\
		\midrule
			Oracle  & 99.5 & 99.4 & 99.3  \\
			\bottomrule
		\end{tabular}
			\caption{Comparison of the performance (\%) for unsupervised domain adaptation. S, M and U are the abbreviations of SVHN, MNIST and USPS respectively. Oracle denotes the target samples are trained in a fully supervised manner.}\label{table3}
	\end{table}

	\begin{figure*}[t]
		\centering
		\subfigure[Source Only]{
			\begin{minipage}[t]{0.3\linewidth}
				\centering
				\includegraphics[width=1.9in, height=1.5in, trim= 0 10 0 0, clip]{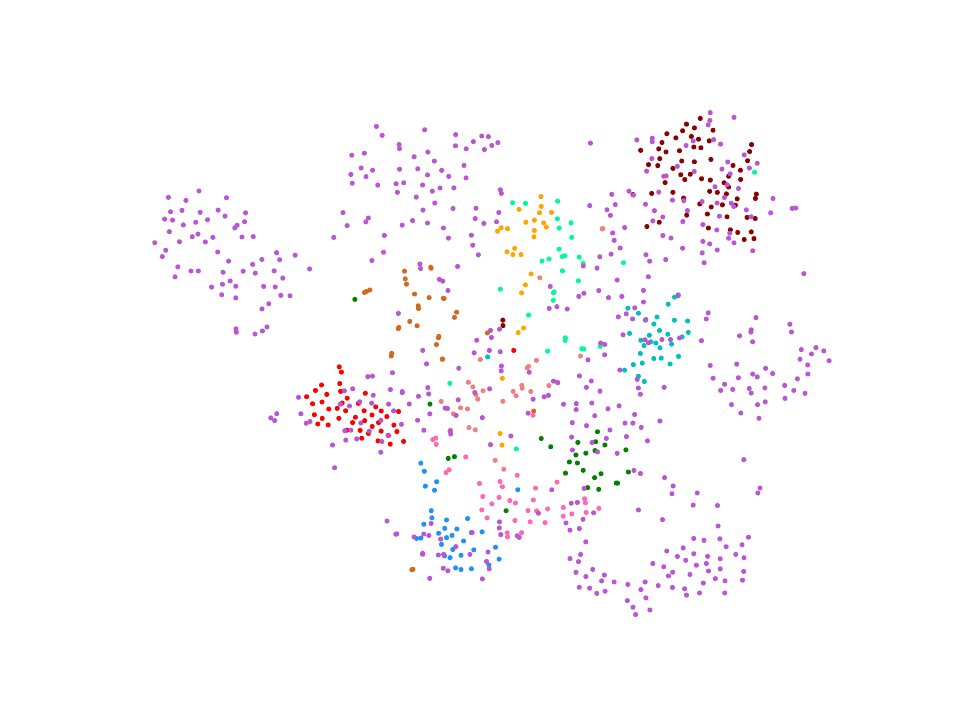}
				%\caption{Source only model}
				\label{figure3a}
			\end{minipage}%
		}%			
		\subfigure[DANN]{
			\begin{minipage}[t]{0.3\linewidth}
				\centering
				\includegraphics[width=1.9in, height=1.5in, trim= 0 10 0 0, clip]{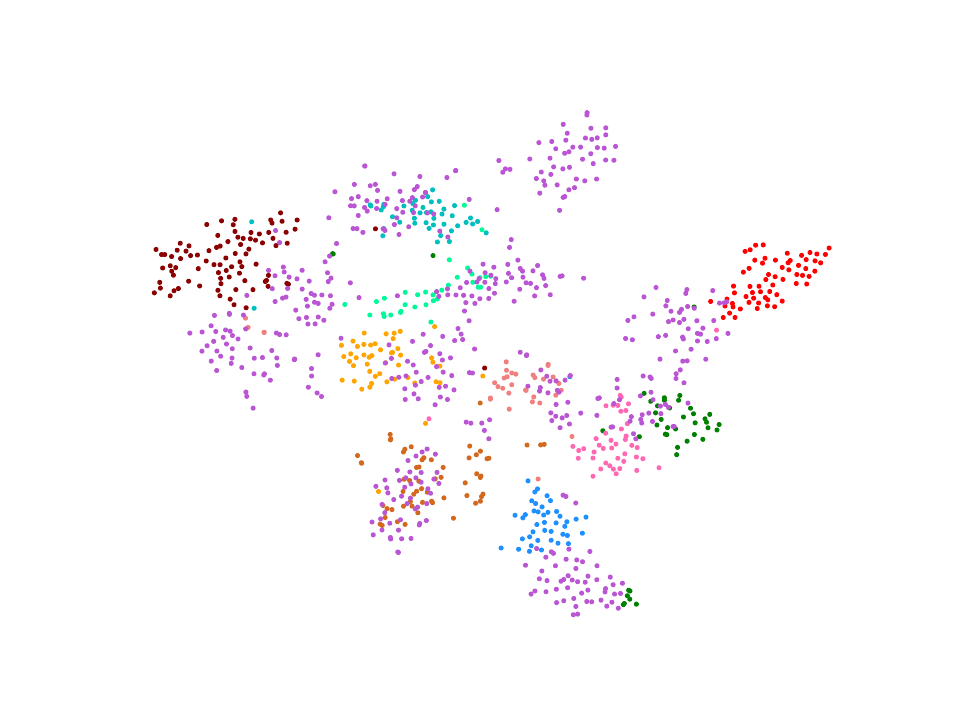}
				%\caption{Source only model}
				\label{figure3b}
			\end{minipage}%
		}%
		\subfigure[Our model]{
			\begin{minipage}[t]{0.3\linewidth}
				\centering
				\includegraphics[width=1.9in, height=1.5in, trim= 0 0 0 0, clip]{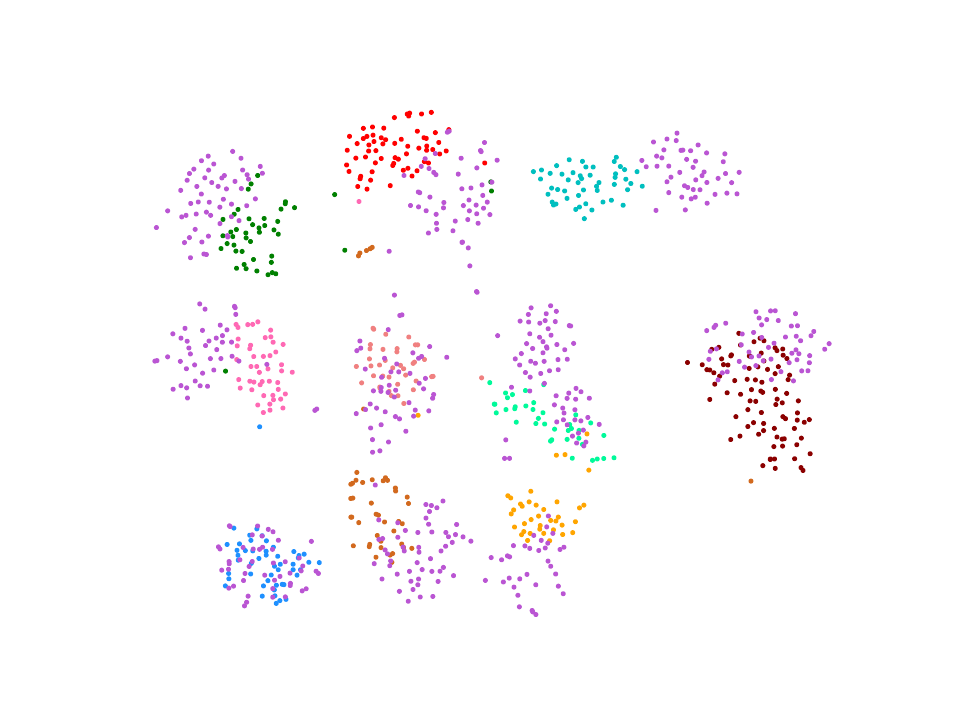}
				%\caption{Our model}
				\label{figure3c}
			\end{minipage}%
		}%
%		\subfigure[ P $\rightarrow$ I]{
%			\begin{minipage}[t]{0.25\linewidth}
%				\centering
%				\includegraphics[width=1.43in, height=1.5in]{p2i_.png}
%				%\caption{P $\rightarrow$ I}
%				\label{figure3c}
%				\vspace{0.38cm}
%			\end{minipage}
%		}%
%		\subfigure[ P $\rightarrow$ C]{
%			\begin{minipage}[t]{0.25\linewidth}
%				\centering
%				\includegraphics[width=1.4in, height=1.5in]{p2c_.png}
%				%\caption{P $\rightarrow$ C}
%				\label{figure3d}
%				\vspace{0.38cm}
%			\end{minipage}
%		}%
		\centering
		\caption{  The feature distribution of samples in two domains are visualized by t-SNE in the task MNIST $\rightarrow$ USPS. Source features are marked in violet, and target features marked in other different colors represent different categories. The feature representations are scattered irregularly between domains learned by the Source Only model. Classic adversarial-based method DANN only take account of global feature alignment in the level of domain. By contrast, our model consider the category-level alignment. The feature representations that belongs to the same category but in different domains are close. Hence, the decision boundary of task-specific classifier can be easily learned.}
		\label{figure3}
	\end{figure*}
	
	\begin{figure*}[htb]
		\centering
		\includegraphics[width=0.95\linewidth]{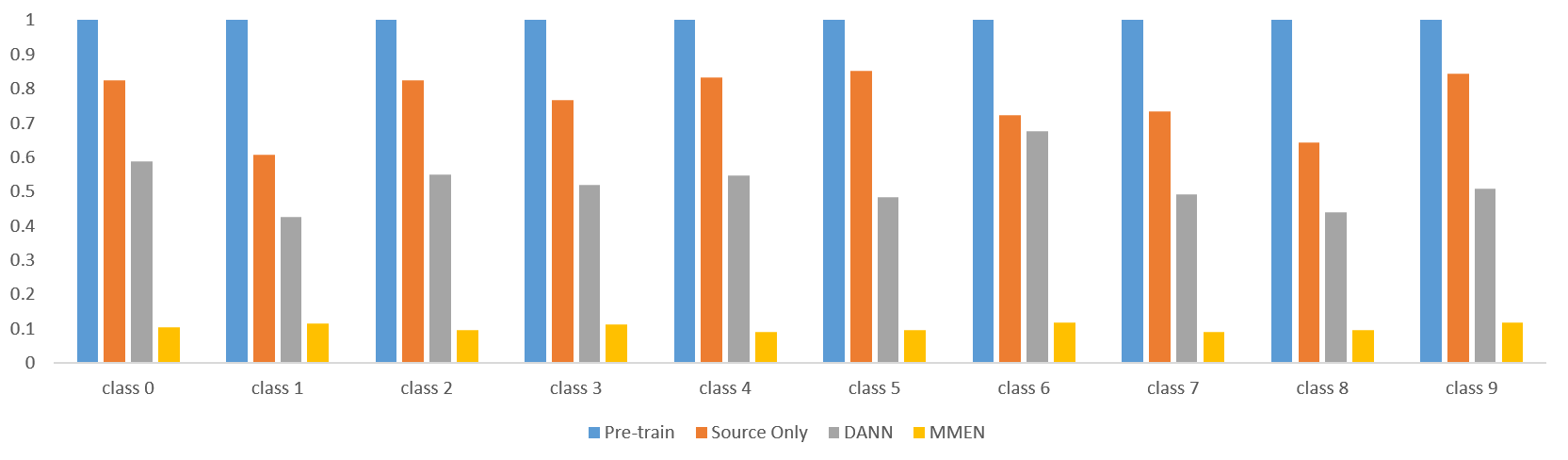}
		\caption{ Analysis of the category-level feature alignment in our model. For each category, we computer the distance of the feature center between source domain and target domain in the task MNIST $\rightarrow$ USPS. Compared with the performance learned by  model pre-trained on the ImageNet, source only model and classic adversarial-based method DANN, our model successfully aligns the feature distribution in all categories. }
		\label{figure5}
	\end{figure*}

	\begin{figure}
			\subfigure[ P $\rightarrow$ I]{
				\begin{minipage}[t]{0.5\linewidth}
					\centering
					\includegraphics[width=1.43in, height=1.5in]{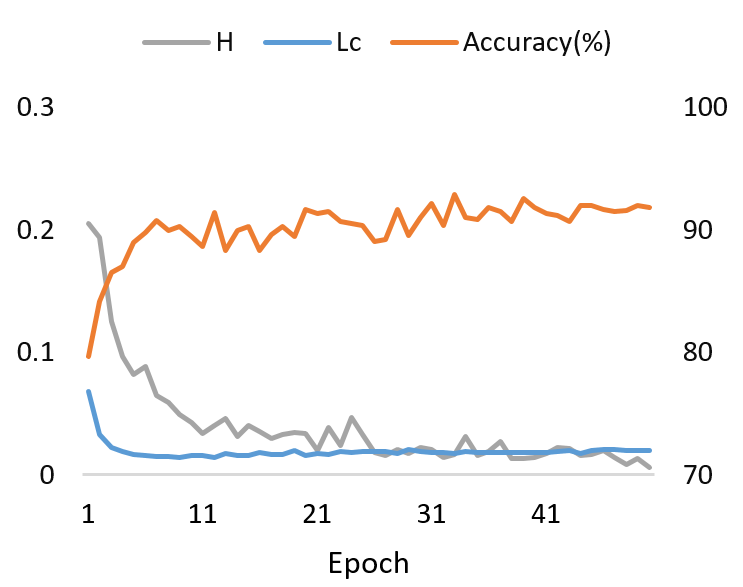}
					%\caption{P $\rightarrow$ I}
					\label{figure4a}
					\vspace{0.38cm}
				\end{minipage}
			}%
			\subfigure[ P $\rightarrow$ C]{
				\begin{minipage}[t]{0.5\linewidth}
					\centering
					\includegraphics[width=1.4in, height=1.5in]{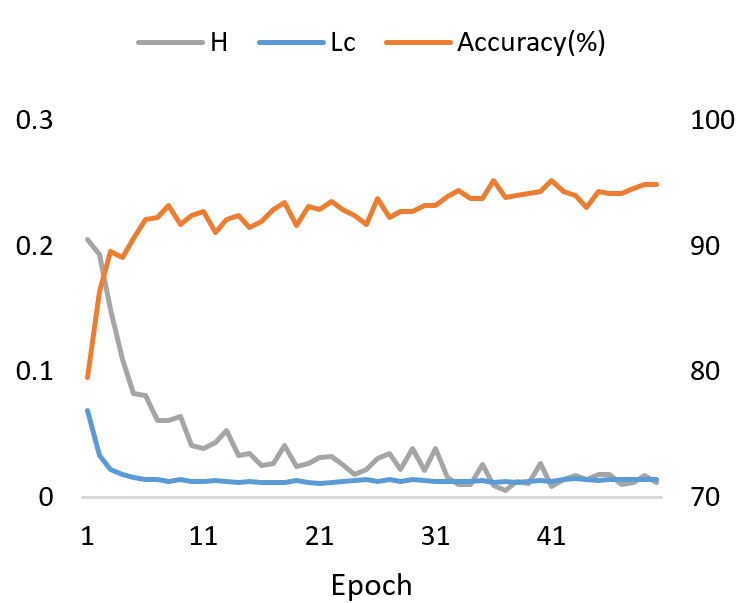}
					%\caption{P $\rightarrow$ C}
					\label{figure4b}
					\vspace{0.38cm}
				\end{minipage}
			}%
		\caption{ The visualization of accuracy and training loss in the task P $\rightarrow$ I and P $\rightarrow$ C on the CLEF datasets. $H$ and $L_c$ denote cross-entropy of pseudo-labels  and cross-entropy with true labels respectively. $Accuracy$ denotes the performance on the classifier.}
		\label{figure4}
	\end{figure}

	\section{Discussion}
	We use the activation of the last layer of generator as feature representations and project them by t-SNE embedding~\cite{maaten2008visualizing}. As shown in the Figure~\ref{figure3a}, we plot the feature distribution learned by source only model. Source features (violet) are distributed dispersedly with categories, which means the classifier can learn the decision boundary for source samples easily. However, target features (other different colors) are scattered irregularly with high ambiguity.  In Figure \ref{figure3b}, since domain classifier only care about the origin of input but ignore categorical information, the generator is only able to align feature in the level of domain globally. Hence, the classifier cannot classify target samples precisely. Contrast with Figure~\ref{figure3c}, target feature are aligned closely with source feature that share the same category but away from those have different categories. It indicates MMEN is able to generate category-invariant feature and enjoy categorical discriminability .
	
	From Figure~\ref{figure4a} and~\ref{figure4b}, we observe that cross-entropy of pseudo-labels   $H(p(x_t))$ declines continuously with the decrease of cross-entropy $L_c$ between $\hat{y_t}$ and  target labels. We only use the target labels in this part for verification, which are not used during training. The former means that our model becomes more and more certain about target labels assignment. The latter indicates that our model successfully infers labels from unlabeled target samples. The simultaneous learning trends of $H(p(x_t))$ and $L_c$ verifies the feasibility of our model. Because of the property of unfairness, the   category discriminator is uncertain about the category of target samples purposely but classify source samples correctly. To minimize the uncertainty of the prediction on target samples, the generator has to mimic the source distribution of specific category in the feature space. As the generated target features become more and more similar with source feature for each category, they can enjoy the classifier's power to classify source samples. Hence, cross-entropy $L_c$ on target samples decreases consecutively during the learning process.

	\begin{table}[t]
		\centering
		\renewcommand\tabcolsep{4.5pt}
		
		\begin{tabular}{lrrrrrrrrr}
			\toprule
			\textbf{$\lambda$} & 0.01 & 0.02    &0.05 &  0.1     & 0.2  & 0.5   &1     &2  \\
			\midrule
			k=2                &96.9 &96.9 &96.8 &96.6 &96.7 &96.5 &93.5 &83.6 \\
			\midrule
			k=3              &  97.6	& 97.4	& 96.9	& 97.2	& 97.7	& 97.4	& 93.7	& 91.6\\
			\midrule
			k=4              & 97.9  & 97.7	& 97.7	& 97.8	& 97.8 & 97.6  & 96.8 & 94.6\\
			\midrule
			k=5              &98.1	&98.4	&98.0	&97.8	&97.8	&97.8	&96.9	&94.8 \\
			\bottomrule
		\end{tabular}
		\caption{(MNIST $\rightarrow$ USPS) Performance (\%) for unsupervised domain adaptation varies with $k$ and $\lambda$.}\label{table4}
	\end{table}
	
	In order to evaluate the category-level feature alignment, we compute the cluster center distance (CCD) $\{ d_1^e,d_2^e, ..., d_K^e \}$ of the same category between two domains in the feature space. $d_K^e$ denotes the CCD of category $K$ in the epoch $e$.  Large cluster center distance means that the feature representations are weakly aligned. Euclidean distance is used as the metric.a We choose 1860 source samples and 1860 target samples in the task MNIST $\rightarrow$ USPS. The value are normalized by dividing the CCD in pre-train model for each category. As shown in Figure \ref{figure5}, the CCD in source only model is large . It is understandable due to the fact that source only model does not align feature distribution. Compared with source only model and DANN, the CCD of all categories in the our model are smaller. This result verify that our model can align the feature distribution in the level of category.
	
	In order to study the sensitivity of our approach, we search coefficient $\lambda$ ranged from 0.01 to 2 and $k$ ranged from 2 to 5 respectively in the task  MNIST $\rightarrow$ USPS.  As reported in Table~\ref{table4}, the performance of our model are high and close with each other despite the variation of $\lambda$ and $k$. In addition, the discriminator D gets stronger quickly  when $\lambda$ gets larger or the times of updating generator in the inner loop $k$ gets smaller. Hence, it causes vanishing gradient and incomplete feature alignment. Accordingly,  performance drops slightly in this case. Generally speaking, our model is robust enough against the variation of hyper-parameter $\lambda$  and $k$.

\begin{figure}[t]
		\centering
		\includegraphics[width=1.0\linewidth]{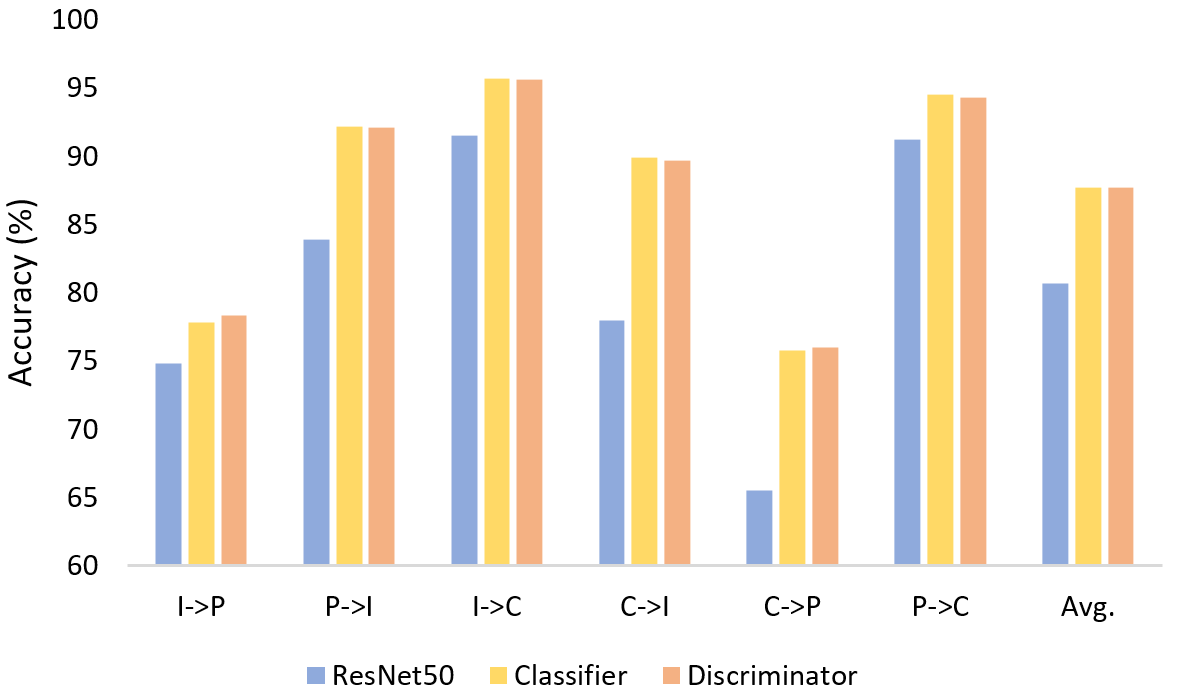}
		\caption{Comparison of performance by ResNet50,  classifier and category discriminator with feature leaned by MMEN. }
		\label{figure6}
	\end{figure}

	Since source risk $\Pr_{(x_s,y_s)\sim D_s}[\eta(x_s)\neq y_s]$ is considerably low after pre-training, we wonder whether a simple classifier can make accurate predictions on unlabeled samples under the premise of well-aligned feature. Therefore, we illustrate the results based on ResNet50 on the ImageCLEF-DA dataset firstly as shown in Figure~\ref{figure6}, and then conduct unsupervised domain adaptations on classifier and category discriminator respectively based on the target feature leaned by MMEN. We observe that both classifier C and discriminator D defeat ResNet50 with large performance gap. Please notice that the classifier is not streamed with target data during training.  The target feature is learned to distributed close with the source one, hence the source classifier C does not need to see the target domain data for good performance. It inspires us that fine-grained feature alignment plays a crucial role in domain adaptation. The well-aligned target features are eligible to utilize the power of source model to infer labels.

	%%%%%%%%% figure %%%%%%%%%%
	%%%%%%%%% figure %%%%%%%%%%
	% & 97.90  & 97.74	& 97.68	& 97.80	& 97.79	& 97.56  & 96.77 & 94.62	& 77.04 & 64.35
	
	\section{Conclusion}
	We have proposed a simple yet effective method for unsupervised domain adaptation. In order to achieve fine-grained aligned feature representations, we inject the target categorical information from target samples directly. Although target labels are unavailable during the learning process, we utilize the cross-entropy of pseudo-labels  to estimate the distribution of class-wise predictions.  By setting an \textit{unfair} category discriminator, we employ adversarial training procedure that push the generative target feature aligned with source feature for each category. Hence, the obtained feature representations enjoy invariance with the shift among categories in different domains. The experimental results demonstrate the superiority of our approach. Moreover, the target feature learned by MMEN exhibits flexible compatibility with source classifiers. 

  \section*{Acknowledgment}

This work was supported by the National Natural Science Foundation of China (No. 62106204) and the Sichuan Natural Science Foundation (No. 2022NSFSC0911).

	\bibliographystyle{named}
	\bibliography{ijcai19}
	
\end{document}